\documentclass{article}

\usepackage[preprint,nonatbib]{neurips_2022}

\usepackage[linesnumbered,ruled,vlined]{algorithm2e}

\usepackage{graphicx}
\usepackage{amsmath}
\usepackage{amsfonts}
\usepackage{caption}
\usepackage[utf8]{inputenc} %
\usepackage[T1]{fontenc}    %
\usepackage{hyperref}       %
\usepackage{url}            %
\usepackage{booktabs}       %
\usepackage{algorithm}            %
\usepackage{nicefrac}       %
\usepackage{microtype}      %
\usepackage{xcolor}         %
\usepackage{multirow}

\title{DAFT: Distilling Adversarially Fine-tuned Models for Better OOD Generalization}

\author{%
  Anshul Nasery$^\dagger$,  Sravanti Addepalli$^\dagger \diamond$, Praneeth Netrapalli$^\dagger$, Prateek Jain$^\dagger$\\
  $^\dagger$ Google Research India  $^\diamond$ Indian Institute of Science, Bangalore\\
  \texttt{\{anshulnasery,sravantia,praneethn,prajain\}@google.com} \\
}
\everypar{\looseness=-1}

\begin{document}
\newcommand{\set}[1]{\left\{#1\right\}}
\newcommand{\Mtilde}{\widetilde{M}}
\newcommand{\cD}{\mathcal{D}}
\newcommand{\alg}{\textsc{DAFT}\xspace}
\newcommand{\algsingle}{\textsc{DAFT-Single}\xspace}
\newcommand{\praneeth}[1]{{\color{red} PN: #1}} %
\newcommand{\pj}[1]{\todo{PJ:#1}} %
\newcommand{\cA}{\mathcal{A}}
\newcommand{\cM}{\mathcal{M}}
\newcommand{\cS}{\mathcal{S}}
\newcommand{\cT}{\mathcal{T}}
\newcommand{\cV}{\mathcal{V}}

\newcommand{\cL}{\mathcal{L}}
\newcommand{\std}[1]{\tiny{$\pm$ #1}}
\newcommand{\stdn}[1]{}
\newcommand{\colormnist}{Color-MNIST}

\maketitle

\begin{abstract}

We consider the problem of OOD generalization, where the goal is to train a model that performs well on test distributions that are different from the training distribution. Deep learning models are known to be fragile to such shifts and can suffer large accuracy drops even for slightly different test distributions  \cite{hendrycks2019benchmarking}.

We propose a new method -- \alg\ --  based on the intuition that adversarially robust combination of a large number of rich features should provide OOD robustness. Our method carefully distills the knowledge \color{black} from a powerful teacher that learns several discriminative features using standard training while combining them using adversarial training. The standard adversarial training procedure is  modified to produce teachers which can guide the student better. We evaluate \alg on  standard benchmarks in the DomainBed framework \cite{gulrajani2020search}, and  demonstrate that \alg achieves significant improvements over the current state-of-the-art OOD generalization methods. \alg consistently out-performs well-tuned ERM and distillation baselines by up to 6\%, with more pronounced gains for smaller networks. 
\end{abstract}

\section{Introduction}
\label{sec:introduction}
Several recent works  have shown that standard deep learning models trained with stochastic gradient descent (SGD) style methods can be {\em fragile} and might suffer a large drop in accuracy if the test data distribution (also known as target domain) is even slightly different compared to the training data distribution (also known as source domain) \cite{hendrycks2019benchmarking, gulrajani2020search}. However, in practice, it is quite challenging to obtain training data that exactly matches the test distribution. For example, due to privacy restrictions we may not be able to access the data of the actual customers of a web-application. Instead, training data is generated using crowd workers or by seeking volunteers who are willing to donate their data for training. This clearly leads to distribution shift between the training and test data. 

Consequently, it is crucial to design models and training mechanisms that are robust to distribution shifts and can perform well on {\em out-of-distribution} (OOD) data. Even in settings where some amount of data can be collected from the final deployment setting, it should be relatively easier to adapt  models with good {\em OOD generalization}.  OOD problems have been studied in a variety of settings where different amounts of source/ target information might be available. We consider the {\em OOD generalization} setting which is one of the weakest settings, and requires only a labeled training dataset, without any information about the target dataset or even about the sources present in the training dataset. Note that this setting is slightly different and more challenging than the popular {\em domain generalization} setting which requires the identity of source domain for each training point. Even, in the domain generalization setting, several recent works \cite{gulrajani2020search, vedantam2021empirical} show that a well-tuned ERM model is still competitive with SOTA methods. %

We propose \alg to mitigate the problem of {\em OOD generalization}; see Section~\ref{sec:problem_definition} for the precise setting. \alg is motivated by three key observations that we discuss below.

Recent works have demonstrated that adversarial training can learn more {\em robust} features compared to standard training \cite{ilyas2019adversarial, yi2021improved}. This in turn can  lead to better performance in few shot and transfer learning settings where final few layers of the network are finetuned using target data \cite{madry2019deep}. 
However, on multiple standard OOD datasets where such finetuning is not feasible, we observe that vanilla  adversarial training does not provide  substantial improvements on larger datasets; see Table~\ref{tab:results}. 

On the other hand, we observe that the representations learned by standard training are already capable of achieving good OOD performance, as is also observed in  \cite{kumar2022fine,rosenfeld2022domain,kirichenko2022last}.
For example, on the iWildCam-WILDS dataset,
if the model is trained with data from a {\em source} domain different from the target domain, the target domain accuracy is around 50\%. However, if we train the model on the {\em source domain} and then \emph{finetune the final layer with a mix of data from the target and the source domain}, the target domain accuracy jumps to 67.1\%, 
demonstrating that features learned by standard training are indeed capable of achieving good accuracy on OOD data. Note that fine-tuning degrades source accuracy by only 2\%, indicating that the features being used are indeed {\em robust}. We conduct further experiments on a binary version of the Coloured-FashionMNIST dataset \cite{xiao2017/fmnist} to establish this intuition that  standard training does learn a few  {\em robust} generalizable features, but they may be drowned out by several non-robust features -- see  Section~\ref{sec:methodology}.

\setlength{\textfloatsep}{3pt}
\begin{figure}[t]
\centering
\includegraphics[width=0.9\linewidth]{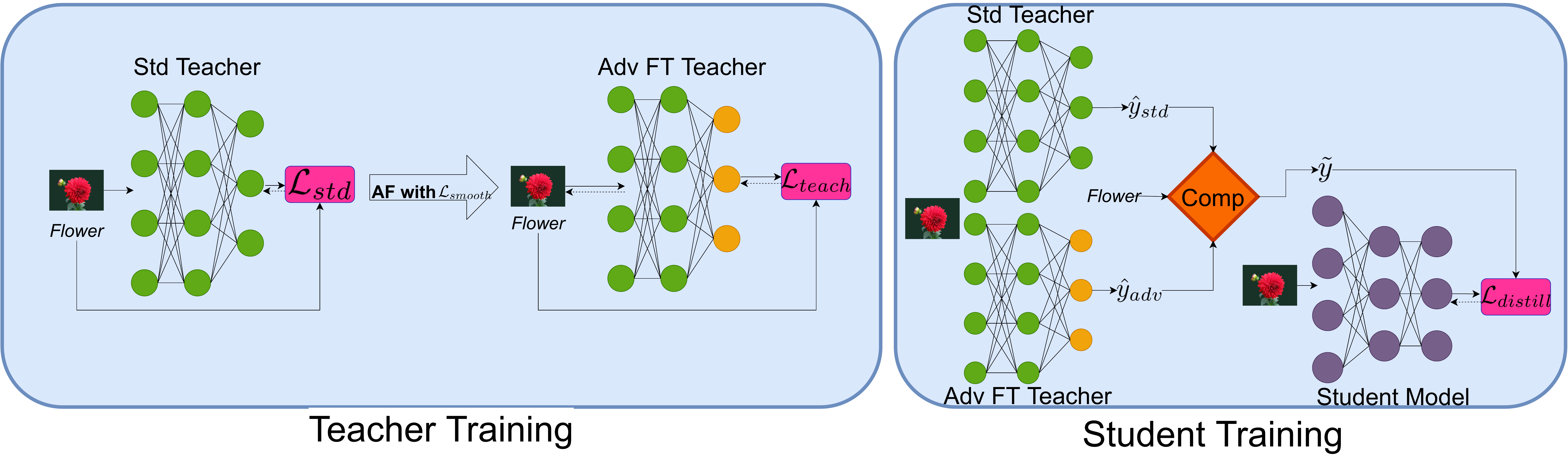}\vspace*{-5pt}
\caption{\alg overview. We pre-train a teacher, followed by adversarial fine-tuning using  $\mathcal{L}_{smooth}$ \eqref{eq:lsmooth}. We then distill a student from both standard and adversarial teachers. The Comp operator outputs $\hat{y}_{adv}$ if adversarial teacher's prediction is correct, else it outputs $\hat{y}_{std}$.
\label{fig:daft}} 
\end{figure}

Motivated by this observation, our first proposal is to pre-train the model using standard training and then, starting from this initialization, perform adversarial training of {\em only} the {\em final linear layer}. We call this \emph{adversarial finetuning}. We observe OOD accuracy gains with this approach itself.

Our second intuition is that, OOD robustness can be improved significantly by diversifying the {\em collection} of standard features.  Distillation~\cite{hinton2015distilling} is known to help a student model access a larger set of ``features" learned by the teacher model. Motivated by this intuition, we studied a method that combines  PGD-AT\cite{madry2019deep} based adversarial finetuning with distillation, i.e., the teacher network is trained with ERM training followed by adversarial finetuning, and then distilled onto a student network. However, this method did not provide significant improvement over just using adversarial finetuning or distilling a standard trained teacher.

An explanation of this behavior is: while adversarial finetuning improves OOD robustness of the teacher, it also makes the logit predictions of all but the top class {\em unstable}. That is, while the ``top" prediction value is stable due to adversarial finetuning loss, the remaining prediction values might change significantly with slight input perturbation. Hence, the predictions are possibly based on an {\em unstable} combination of features, leading to poor OOD performance. To alleviate this concern, we introduce a loss function -- similar to one used by the TRADES algorithm \cite{zhang2019theoretically} -- that penalizes the distance between prediction values with and without adversarial perturbation. This leads to better teacher networks, since the incorrect teacher logits also provide useful information to the student.

Finally, to overcome the challenge of relatively poor accuracy of the adversarially finetuned teacher model on the \emph{training} data distribution, we introduce a way to perform high quality distillation with respect to both the standard and adversarial finetuned teacher models.

\textbf{Our contributions}:
In summary, we introduce a novel method -- Distillation with Adversarially Finetuned Teacher (\alg) -- that uses the above algorithmic insights to design a robust  technique for OOD generalization. \alg trains a student model by distilling with a standard trained as well as an adversarially finetuned teacher model. Finetuning uses  a smooth KL divergence based loss for all logits/prediction values. See Figure~\ref{fig:daft} for an overview of \alg.

We conduct extensive experiments in the standard DomainBed framework \cite{gulrajani2020search} -- using the prescribed methodology for evaluation, hyperparameter tuning --  and  compare \alg against various baselines on the five OOD datasets in the testbed. Recall that even for stronger Domain Generalization setting, \cite{gulrajani2020search} showed that well-trained ERM is nearly SOTA. In contrast, we demonstrate that \alg trained in the weaker OOD generalization setting, still  consistently outperforms  ERM (trained according to DomainBed approach) and other baselines. \alg is particularly effective for smaller network architectures. 
For example, on DomainBed datasets, DAFT trained ResNet-50 models are on an average 4\% more accurate  than ERM as well as other baselines. In fact, DAFT+ResNet-50 models are more accurate than an ERM trained ResNet-152 as well; see Table~\ref{tab:results}. %
Finally, we conducted  ablation studies to analyze the benefit of each of the algorithmic components introduced by \alg. We observe that a combination of all the components of \alg is critical towards achieving overall strong OOD performance. 

\textbf{Limitations}: In general, different application areas/problems might lead to different forms of distribution shift between training and test distributions. Consequently, there may not be a single algorithm that is optimal under all distribution shifts. To address this, we perform experiments on multiple OOD datasets with different characteristics such as satellite images with temporal/geographical shift in TerraIncognita, and object images with varying sources in OfficeHome, VLCS, PACS and DomainNet datasets. In all of these settings, \alg gives significant improvements over ERM. However, there may certainly be other forms of distribution shift, beyond those captured in the above datasets, where new ideas might be required to obtain superior performance.

\section{Related Work}
\label{sec:related_work}
Generalization of a model to new domains has been studied in various contexts such as domain adaptation~\cite{bendavid2007domain}, domain generalization~\cite{gulrajani2020search}, OOD  robustness/generalization~\cite{shen2021towards,hendrycks2019benchmarking} and adversarial robustness~\cite{szegedy2013intriguing,madry2019deep}. We now briefly review prior work from these areas.
\paragraph{Domain generalization and adaptation} In domain generalization, the training data is drawn from multiple domains or environments, each with a different distribution, and each training data point comes with its associated domain label. This fact is exploited by  existing approaches  by: a) learning domain invariant features\cite{arjovsky2019invariant,zhao2019learning}, b) matching feature representations across domains~\cite{sun2016deep, li2018domain}, c) matching gradients across domains~\cite{shankar2018generalizing, shen2021towards}, d) performing distributionally robust optimization across various domains ~\cite{sagawa2019distributionally}, e) augementing data with linearly interpolated samples from different domains~\cite{wang2020heterogeneous, yao2022improving}, f) learning common features across all the domains~\cite{li2017deeper,piratla2020efficient}, and g)  learning causal mechanisms~\cite{scholkopf2012causal}.  However, recent work~\cite{gulrajani2020search} has shown that most of these methods perform no better than a well-tuned ERM over a larger model. Notable exceptions  include results by  \cite{cha2021swad} and a few others mentioned below. %
\cite{cha2021swad} proposes a training method to ensure  flat minima which should lead to robustness.  However, their technique assumes access to domain labels for creating batches and hence does not directly apply to our setting. It is also orthogonal to our work and can potentially be combined if batches are formed randomly. {\em Concurrent and independent} works  in domain generalization setting either fine-tune with large pretrained  models~\cite{kumar2022fine}, or use the features from ERM trained models~\cite{rosenfeld2022domain}, or  transfer features from large models to smaller models~\cite{cha2022miro}. 

Unsupervised domain adaptation is a related  problem setting \cite{ganin2016domainadversarial,courty2016optimal,choi2019pseudo,masashi2007direct,jiang-zhai-2007-instance}, but these techniques rely on access to unlabelled target domain data and hence, do not apply to our setting.%
\paragraph{OOD Robustness/Generalization}  We study this setting, wherein no assumptions are made on the training or the test data domains. Due to a large diversity in   domain shift mechanisms, several benchmarks have been introduced to evaluate OOD generalization including artificially induced domain shifts such as blurring, adding noise etc.~\cite{hendrycks2019benchmarking}, temporal, geographic and demographic shifts in natural settings~\cite{koh2021wilds,malinin2021shifts} etc. 
One approach to tackle this problem is to learn networks which side-step spurious correlations between the inputs and labels, using model pruning techniques~\cite{zhang2021subnetwork}. We show that this method is less accurate than \alg. 

Recently, \cite{miller2021accuracy} demonstrated that larger models lead to better in-domain accuracy, which consequently leads to better OOD generalization. We exploit this intuition to try to transfer better features from larger models to smaller ones via distillation. In fact, using our  method, significantly smaller architectures can  outperform larger architectures in terms of OOD accuracy; see Table~\ref{tab:results}. \vspace*{-10pt}
 \paragraph{Adversarial Robustness}: Output of deep networks is known to be sensitive to small, carefully crafted perturbations of the  input~\cite{szegedy2013intriguing}, and adversarial training (AT)~\cite{madry2019deep} has been proposed to make models robust to such perturbations. A similar line of work, known as distributionally robust optimization trains networks to minimize the worst case performance on distributions in a small ball around the empirical  distribution~\cite{ben2013robust, shafieezadeh2015distributionally, duchi2021learning, levy2020largescale}. %
Another line of work propose modifying the training objective of AT to ensure that \emph{all} logits are similar for the clean and adversarial  example~\cite{kannan2018adversarial, zhang2019theoretically}. There have also been works which use adversarial training to promote adversarial \emph{distributional} robustness in the learnt models~\cite{sinha2017certifying,lee2017minimax,yi2021improved, Zi_2021_ICCV}, bringing it closer to OOD robustness. However, the bounds in these works break down when the distribution shift is large, e.g. for datasets like PACS or VLCS.
In this work, we demonstrate that, for OOD robustness, adversarial training of last layer tends to significantly outperform  adversarial training of the entire network. While we use projected gradient descent (PGD) \cite{madry2019deep} and TRADES \cite{zhang2019theoretically} methods to demonstrate the above claim due to their good empirical performance, we can in principle use any other AT method in their place.

\paragraph{Knowledge Distillation}:
Knowledge distillation was introduced as a model compression technique which teaches a smaller model to mimic the outputs of a larger teacher~\cite{hinton2015distilling}. Related works have also attempted at making students more robust to adversarial perturbations~\cite{goldblum2020adversarially,zhu2021reliable} as well as using intermediate teacher representations to guide the student~\cite{tian2019contrastive} but such methods do not consider OOD robustness of students. Furthermore, we demonstrate in Section~\ref{sec:experimental_results} that \alg outperforms vanilla distillation and distillation from an adversarially finetuned teacher (Table~\ref{tab:results}, Fig~\ref{fig:smoothness}).

In summary, we find that most OOD works assume access to  domain labels in one form or the other. In the OOD generalization setting where this assumption is dropped, ERM performs the best \cite{gulrajani2020search}. We show that DAFT consistently outperforms ERM,  especially  for smaller models.

\vspace*{-2pt}
\section{Problem Definition}\vspace*{-4pt}
\label{sec:problem_definition}

\textbf{Problem setting}: We are given a training dataset of labeled  examples $\cD_S = \set{(x_i,y_i):i\in [n]}$, comprising of images from a single or multiple distributions. We do not assume access to domain labels. We assume access to a validation split from the test domain $\cT$, which could be used for model selection in principle. In order to prevent the selection of hyperparameters that are tuned specifically to the target domain, we utilize the leave-one-domain-out validation method suggested by Gulrajani et al. \cite{gulrajani2020search}. This ensures that a common set of hyperparameters can be used for any combination of source domains in the dataset. While this was used for the purpose of evaluation and benchmarking, in practice, the target domain split alone could be used for model selection, which would lead to improvements over the reported results. We would like to emphasize that we do not have access to $\cT$ during model training. 
While we do not assume any relationship between the training and the test data distributions, the datasets under consideration exhibit covariate shift, wherein the data marginal changes, while the conditional label distribution does not. Hence, we study this setting as it is more practically applicable.
\color{black}

\textbf{Motivation}: In several settings, there are privacy and proprietary reasons for not having access to the validation or target data (i.e., $\cV$ and $\cT$ respectively) during model training. 
However, once we train a model, we often have the ability to deploy/ evaluate the model for its accuracy on the validation domain $\cV$, and in some cases, even on the target domain $\cT$.

\vspace*{-2pt}
\section{Method -- \alg}\vspace*{-4pt}
\label{sec:methodology}
\begin{algorithm*}[t]
    \SetAlgoLined
    \DontPrintSemicolon
	\KwData{Training data $\cD_S = \set{(x_i,y_i):i\in [n]}$, teacher model $(\theta,W)$, student model $(\mu,U)$, $\epsilon$, $\alpha$, Boolean: $\textrm{SingleDistill}$.}
	$\theta_\textrm{std}$, $W_\textrm{std} \leftarrow \textrm{Adam}\left(\min_{\theta, W} \sum_i \cL_{\textrm{std}}(\theta, W, (x_i, y_i))\right)$. \\
	\tcc*{Standard training of teacher model parameters $\theta$ and $W$.}
	
	Using $\theta_\textrm{std}$, $W_\textrm{std}$ as initialization, optimize only $W$: \\
	$W_{\textrm{adv}} \leftarrow \textrm{Adam}\left(\sum_i \cL_{\textrm{adv},\epsilon,\theta}(W, (x_i, y_i)) + \alpha \cdot \cL_{\textrm{smooth},\epsilon,\theta}(W, (x_i,y_i))\right)$. \\
	\tcc*{{Smooth adversarial finetuning of final linear layer $W$.}}

	\eIf(\tcc*[h]{If $\textrm{SingleDistill}$ is set to $\textrm{True}$ then}){\textrm{SingleDistill}}{\algsingle: ${\mu^*, U^*} = \textrm{Adam}\left( \sum_i \cL_{\textrm{distill}}(\mu, U, (x_i, \tilde{y}_i))\right)$, \\
	where $\tilde{y}_i =\frac{ \exp\left(W_\textrm{adv}^\top f_{\theta_\textrm{std}}({x_i})/\tau\right)}{\sum_{\hat{y}}\exp\left(w_{_\textrm{adv},\hat{y}}^\top f_{\theta_\textrm{std}}({x_i})/\tau\right)}$ \\ \tcc*{Train the student model using distillation loss $\cL_{\textrm{distill}}$ defined in Eqn. \eqref{eqn:distill}, using only the adversarially finetuned teacher $(\theta_\textrm{std},W_\textrm{adv})$.}}{	\alg: ${\mu^*, U^*} = \textrm{Adam}\left(\sum_i \cL_{\textrm{distill}}(\mu, U, (x_i, \tilde{y}_i))\right)$, \\
	where $\tilde{y}_i = \exp\left(W_\textrm{adv}^\top f_{\theta_\textrm{std}}({x_i})/\tau\right)/\sum_{\hat{y}}\exp\left(w_{_\textrm{adv},\hat{y}}^\top f_{\theta_\textrm{std}}({x_i})/\tau\right)$ if $(\theta_\textrm{std},W_\textrm{adv})$ correctly outputs $y_i$; else $\tilde{y}_i = \exp\left(W_\textrm{std}^\top f_{\theta_\textrm{std}}({x_i})/\tau\right)/\sum_{\hat{y}}\exp\left(w_{_\textrm{adv},\hat{y}}^\top f_{\theta_\textrm{std}}({x_i})/\tau\right)$. \\ \tcc*{Train the student model using distillation loss $\cL_{\textrm{distill}}$  \eqref{eqn:distill}, where teacher logits are computed using adversarially finetuned teacher if it predicts correctly on this example; standard  teacher otherwise.}}

	\KwResult{Trained model $(\mu^*, U^*)$.}

	\caption{DAFT: Distillation of Adversarially Fine-tuned Teacher}
	\label{alg:alg}
\end{algorithm*}

\textbf{Motivation and high level description of algorithm}:
We begin by describing the key insights that led to our algorithm, and provide a high-level description. First key observation is that standard ERM trained models might already learn features which are good for domain generalization \cite{kirichenko2022last, kumar2022fine, rosenfeld2022domain}, but the final layer is not able to combine these features in a manner robust to domain shifts.  

To further illustrate this point, we perform an experiment using a modification of a Fashion-MNIST subset with only two-class: \textit{shoe} and \textit{top}. We superimpose images onto coloured backgrounds, where the colour varies linearly between red $(255, 0, 0)$ and green$(0, 255, 0)$. Training images have a strong correlation between the background colour and label, i.e. color of \textit{tops} images range from $(255,0,0)$ to $(123,132,0)$, while that of  \textit{shoes} images range between $(132,123,0)$ and $(0,255,0)$. In general, color can easily distinguish between the classes, but there is a small region between $(123,132,0)$ and $(132,123,0)$, where color cannot distinguish between the classes. During test time, there is no such correlation with colour i.e. the data is OOD w.r.t. the train data. For models trained on this data, we compute the correlation of each neuron at the output of the feature extractor with the shape and colour of the images. Note that color is a non-robust spurious feature while shape is a robust feature that is strongly correlated with the labels and is useful for prediction despite OOD shifts.

Now, an ERM trained model has an in-domain (ID) test  accuracy of 99.9\%, and OOD test accuracy of 60\%. Furthermore, {\em only} 2 of the 32 features are highly  correlated with the robust shape feature, while the rest are correlated with color. The final class output is dominated by the color features. %
In contrast, an adversarially trained model has a higher  number of shape features (8 out of 32). But the in-domain accuracy is only 98.3\%, and the OOD accuracy  is 58\%, lower than ERM. A probable reason is that even though adversarial training learns \emph{more} shape-correlated robust features, but the average correlation with shape is much smaller (around $0.75$) than the similar correlation of features from standard trained model (around $0.8$). This is possibly because the shape features learned by adversarially trained models are more suited to the goal of adversarial robustness, while the features learned by standard ERM models are better correlated with the standard classification task.
Hence, we introduce the method of adversarial fine-tuning of the \emph{last layer} after standard ERM based pre-training.  This encourages the model to give a lower prediction weight to color features, and higher weight to the robust shape features learned by the ERM model. With adversarial fine-tuning, a small ID accuracy drop 99.5\% occurs, but the OOD accuracy jumps to 64\%. 

Next, motivated by the observation that distillation from larger models often helps in-domain performance, we trained our final model through distillation of a larger model which was itself trained using adversarial fine-tuning.

Surprisingly, vanilla distillation failed to transfer the superior performance of teacher model to the student model. We identify the main reason behind the failure of distillation in transfering the superior OOD performance of teacher to the student to be the following: while the Cross-Entropy loss on adversarial samples ensures that the logit corresponding to the correct class is ``robust", it does not put any constraints on the remaining logits. Consequently, the remaining logits do not provide useful information for distillation. In order to tackle this, we add an additional KL divergence term to ensure that all the logits of the teacher model are smooth in the neighborhood of the given input, and are aligned to the logits of the clean image.

The final ingredient of our approach is to use two teacher models: the adversarially fine-tuned teacher on inputs where it predicts correctly and the standard trained model on the remaining inputs. We call the resulting algorithm Distillation of Adversarially Fine-Tuned teacher (\alg). We now present each of the components in \alg in more detail.

\textbf{Adversarial fine-tuning}: The first component is adversarial fine-tuning, where we first use standard training and then, using this as a pre-trained initialization, perform adversarial training of the final linear layer. The loss functions used in the standard pre-training and adversarial fine-tuning on a given data point $(x,y)$ are $\cL_\textrm{std}(\theta, W, (x,y))$ and $\cL_{\textrm{adv},\epsilon,\theta}(W, (x,y))$, respectively where, 
\begin{align*}
\resizebox{0.98\hsize}{!}{
    $\cL_{\textrm{std}}(\theta,W,(x,y)) = \log \frac{\sum_{\hat{y}} \exp\left(w_{\hat{y}}^\top f_\theta({x})\right)}{\exp\left(w_{{y}}^\top f_\theta({x})\right)},\ 
    \cL_{\textrm{adv},\epsilon,\theta}(W,(x,y)) = \max_{\hat{x} \in B_\epsilon(x)} \log \frac{\sum_{\hat{y}} \exp\left(w_{\hat{y}}^\top f_\theta(\hat{x})\right)}{\exp\left(w_{{y}}^\top f_\theta(\hat{x})\right)},$
    }
\end{align*}
where $f_{\theta}(x)$ denotes the penultimate layer representation of $x$ and $W$ is the final linear layer. \\
Note that while previous studies have shown that adversarial perturbations applied to the input of a network are amplified in the logit space, we tune the value of $\epsilon$ for getting the best performance on validation OOD data, which means that we can side-step this issue by having smaller perturbations. Further, we posit that while the perturbations in feature space will be large, the features which will be perturbed more are the non-robust features. We verify this empirically on the coloured-FashionMNIST dataset in the appendix, where we find that features corresponding to color get perturbed more than shape features. We would want our model to ignore such features. The training objective would ensure that this happens, leading to more robust models. We also conduct experiments with perturbing examples in the feature space and find that both these approaches yield similar empirical OOD performance (see Appendix). 

\textbf{Distillation from a larger teacher model}: The second component is logit distillation -- instead of training the model directly on the training dataset, we first train a larger teacher model using adversarial fine-tuning on the training dataset and then use logit distillation to train the desired model. Given an input $x$, temperature $\tau$ and teacher model $W$, the distillation loss for the student is given by:
\begin{align}
    \cL_{\textrm{distill}}(\mu, U, (x, \tilde{y})) = KL(z, \tilde{y}),\label{eqn:distill}
\end{align}
$z=\exp\left(U^\top f_\mu(x)/\tau\right)/\sum_{\hat{y}} \exp\left(u_{\hat{y}}^\top f_\mu(x) / \tau\right)$, $\tilde{y} = \exp\left(W^\top f_\theta(x)/\tau\right)/\sum_{\hat{y}} \exp\left(w_{\hat{y}}^\top f_\theta(x)/\tau\right)$. Here $f_{\mu}(x)$ denotes the penultimate layer representation of $x$ and $U$ is the final linear layer corresponding to the student model, while $\theta$\color{black} and $W$ are the corresponding parameters of the teacher.

\textbf{Smoothness of incorrect logits}: The cross entropy loss function $\cL_{\textrm{adv},\epsilon,\theta}$ in adversarial fine-tuning of teacher ensures that the logit corresponding to the correct class is large but does not explicitly consider logits corresponding to incorrect classes. However, logit distillation uses logits of the teacher model corresponding to all classes to train the student model. In order to ensure that logits of the teacher model corresponding to all classes are meaningful, we add an additional KL divergence term between logits of the given input and any other point in an $l_2$ norm constrained ball $B_\epsilon(x)$ \color{black} around it.
\begin{equation}
    \cL_{\textrm{smooth},\epsilon,\theta}(W,(x,y)) = \max_{\hat{x} \in B_\epsilon(x)} KL(z||\hat{z}),\label{eq:lsmooth}
\end{equation}
where $z=\exp\left(W^\top f_\theta({x})\right)/\sum_{\hat{y}}\exp\left(w_{\hat{y}}^\top f_\theta({x})\right)$ and $\hat{z}=\exp\left(W^\top f_\theta(\hat{x})\right)/\sum_{\hat{y}}\exp\left(w_{\hat{y}}^\top f_\theta(\hat{x})\right)$. So the loss function we use in adversarial fine-tuning of the teacher is $\cL_{\textrm{adv},\epsilon,\theta} + \alpha \cdot \cL_{\textrm{smooth},\epsilon,\theta}$ for some $\alpha>0$. We note that this cumulative loss function is similar to that in TRADES with the key difference being that TRADES uses $\cL_{\textrm{std},\theta}$ in addition to the KL divergence term, while we use $\cL_{\textrm{adv},\epsilon,\theta}$ term. We use $l_2$ constrained perturbations, with the radius $\epsilon$ being a hyper-paremeter (whose value is mentioned in the appendix for various datasets).

\textbf{Multi-distillation}: Finally, as adversarial fine-tuning hurts in-domain accuracy and predicts incorrect labels for some in-domain training examples, we use different teachers for computing teacher logits on different examples. For those examples on which adversarially fine-tuned teacher predicts the correct label, we use the logits of adversarially fine-tuned teacher, while for those examples on which it predicts incorrectly, we use the logits of the standard trained teacher. More concretely, the loss function for \alg is:
$\min_{\mu, U} \sum_i \cL_{\textrm{distill}}(\mu, U, (x_i, \tilde{y}_i))$,
where $\tilde{y}_i = \exp\left(W_\textrm{adv}^\top f_{\theta_\textrm{std}}({x_i})/\tau\right)/\sum_{\hat{y}}\exp\left(w_{_\textrm{adv},\hat{y}}^\top f_{\theta_\textrm{std}}({x_i})/\tau\right)$ if $(\theta_\textrm{std},W_\textrm{adv})$ correctly outputs $y_i$. Else, $\tilde{y}_i = \exp\left(W_\textrm{std}^\top f_{\theta_\textrm{std}}({x_i})/\tau\right)/\sum_{\hat{y}}\exp\left(w_{_\textrm{adv},\hat{y}}^\top f_{\theta_\textrm{std}}({x_i})/\tau\right)$. Here, $(\theta_{\textrm{std}},W_\textrm{adv})$ is the adversarially fine-tuned model, $(\theta_{\textrm{std}},W_\textrm{std})$ is the standard trained model. For ablation studies, we also present results for the vanilla version of \alg, called \algsingle, where we only use the adversarilly finetuned teacher. The loss function for \algsingle is the same as above, except that $\tilde{y}_i = \exp\left(W_\textrm{adv}^\top f_{\theta_\textrm{std}}({x_i})/\tau\right)/\sum_{\hat{y}}\exp\left(w_{_\textrm{adv},\hat{y}}^\top f_{\theta_\textrm{std}}({x_i})/\tau\right)$ for all $i$.
A pseudocode for the full algorithm including all these components is presented in Algorithm~\ref{alg:alg}. 

\section{Experimental Results}\vspace*{-4pt}
\label{sec:experimental_results}
\begin{table}
\begin{center}
\begin{small}
\begin{sc}
\resizebox{\linewidth}{!}{
\begin{tabular}{c|cccccc|c}
\toprule
Model Size                  & Method               & PACS & VLCS & OfficeHome & DomainNet  & TerraIncognita  & Avg \\
\midrule
\multirow{5}{*}{ResNet-18} & ERM & 80.2\std{1.0} & 71.4\std{0.6} & 57.4\std{0.4} & 31.2\std{0.0} & 40.8\std{1.3} & 56.2\\
 & AT & 79.6\std{0.9} & 68.6\std{0.3} & 56.5\std{0.8} & 30.6\std{0.6} & 56.5\std{0.7} & 58.4\\
 & TRADES & 79.4\std{0.6} & 70.4\std{0.8} & 56.7\std{0.7} & 29.8\std{0.1} & 39.6\std{0.9} & 55.2\\
 & Distillation & \textbf{83.1\std{0.3}} & 76.6\std{0.3} & 62.8\std{0.3} & 33.8\std{0.2} & 48.3\std{0.5} & 60.9\\
 & DAFT & \textbf{84.7\std{1.1}} & \textbf{78.2\std{0.1}} & \textbf{63.2\std{0.2}} & \textbf{36.4\std{0.2}} & \textbf{50.2\std{0.8}} & \textbf{62.5}\\
\midrule
\multirow{5}{*}{ResNet-34} & ERM & 83.2\std{0.8} & 73.5\std{1.0} & 60.8\std{0.6} & 32.5\std{0.0} & 41.0\std{0.7} & 58.2\\
 & AT & 82.2\std{1.0} & 72.9\std{0.5} & 60.5\std{0.5} & 30.7\std{0.3} & 40.6\std{0.4} & 57.4\\
 & TRADES & 82.5\std{0.6} & 72.2\std{0.7} & 60.7\std{0.8} & 31.4\std{0.3} & 41.3\std{0.2} & 57.6\\
 & Distillation & 84.0\std{1.7} & 76.0\std{0.7} & 66.3\std{0.1} & 36.7\std{0.1} & 48.5\std{0.9} & 62.3\\
 & DAFT & \textbf{87.4\std{0.3}} & \textbf{79.1\std{0.9}} & \textbf{67.2\std{0.5}} & \textbf{38.5\std{0.3}} & \textbf{51.4\std{1.1}} & \textbf{64.7}\\
\midrule
\multirow{5}{*}{ResNet-50} & ERM & 83.3\std{1.7} & 75.2\std{1.2} & 67.0\std{0.6} & 41.1\std{0.1} & 46.2\std{0.7} & 62.6\\
 & AT & 82.6\std{1.2} & 72.0\std{1.2} & 67.0\std{0.3} & 40.3\std{0.2} & 45.3\std{1.1} & 61.4\\
 & TRADES & 82.6\std{0.9} & 72.3\std{0.9} & 66.1\std{0.8} & 40.4\std{0.1} & 45.1\std{0.4} & 61.3\\
 & Distillation & 85.9\std{0.9} & 76.5\std{0.9} & 67.7\std{0.4} & 41.9\std{0.2} & 50.7\std{0.7} & 64.5\\
 & DAFT & \textbf{88.0\std{0.1}} & \textbf{80.0\std{0.2}} & \textbf{71.0\std{0.2}} & \textbf{42.6\std{0.2}} & \textbf{52.8\std{0.1}} & \textbf{66.9}\\
\midrule
\multirow{5}{*}{ResNet-101} & ERM & 85.0\std{0.0} & 76.9\std{0.4} & 67.6\std{0.5} & 42.6\std{0.1} & 49.5\std{0.0} & 64.3\\
 & AT & 72.6\std{0.1} & 75.9\std{0.4} & 67.5\std{0.4} & 42.3\std{0.1} & 47.9\std{0.1} & 61.2\\
 & TRADES & 83.7\std{0.5} & 76.3\std{0.4} & 68.0\std{0.3} & 42.2\std{0.1} & 49.5\std{0.8} & 63.9\\
 & Distillation & 86.9\std{0.7} & 77.1\std{0.4} & 69.1\std{0.2} & \textbf{43.2\std{0.1}} & 50.3\std{0.3} & 65.3\\
 & DAFT & \textbf{88.8\std{0.5}} & \textbf{79.1\std{0.5}} & \textbf{72.2\std{0.8}} & \textbf{43.7\std{0.5}} & \textbf{54.1\std{0.9}} & \textbf{67.6}\\
\midrule
\multirow{5}{*}{ResNet-152} & ERM & 87.0\std{0.4} & 79.2\std{0.1} & 69.0\std{0.5} & 43.2\std{0.0} & 50.4\std{0.2} & 65.7\\
 & AT & 87.1\std{0.1} & 78.8\std{0.1} & 69.6\std{0.3} & 42.8\std{0.0} & 49.6\std{0.5} & 65.6\\
 & TRADES & 87.3\std{0.1} & 78.8\std{0.1} & 69.7\std{0.1} & 42.7\std{0.0} & 49.8\std{0.2} & 65.7\\
 & Distillation & \textbf{88.8\std{1.6}} & \textbf{80.4\std{1.3}} & \textbf{71.3\std{0.2}} & 43.6\std{0.1} & \textbf{55.1\std{1.2}} & 67.8\\
 & DAFT & \textbf{88.7\std{2.0}} & \textbf{80.7\std{1.7}} & \textbf{71.9\std{1.2}} & \textbf{44.1\std{0.0}} & \textbf{55.9\std{1.0}} & \textbf{68.3}\\

\bottomrule
\end{tabular}}\vspace*{2pt}
\caption{OOD accuracy on various datasets with different ResNet (RN) architectures.
\label{tab:results}
}
\end{sc}
\end{small}
\end{center}

\end{table}
In this section, we detail our experimental setup, datasets, baselines and results.
\vspace*{-5pt}
\subsection{Experimental Setup}
Our experimental setup follows the approach and recommendations of \cite{gulrajani2020search}. 
For all our experiments, we train models of different sizes from the ResNet~\cite{he2015deep} family. We also perform augmentations including random cropping, flipping and colour jitter on the training data. Hyper-parameters for all the methods are tuned using the \textit{leave-one-domain-out} approach descibed in \cite{gulrajani2020search}. 
See Appendix for additional details on hyperparameters.  We use ImageNet pretrained models for comparisons on DomainBed.
We report the mean and std deviation of the metrics across five random restarts.

\vspace*{-6pt}
\paragraph{Datasets}
We report OOD accuracy results on 5 different datasets, and the average accuracy across them. We use all the datasets from the DomainBed~\cite{gulrajani2020search} benchmark (i.e. PACS\cite{li2017deeper}, VLCS\cite{fang2013unbiased}, OfficeHome\cite{venkateswara2017deep}, DomainNet\cite{peng2019moment} and TerraIncognita\cite{beery2018recognition}),
whose details are in the appendix.

\vspace*{-5pt}
\paragraph{Baselines}
We compare \alg against the standard ERM method  trained on training data $\cD_S$. We also compare against AT which trains models using PGD for $l_2$-norm constrained input  adversarial perturbations \cite{madry2019deep}, as well as TRADES \cite{zhang2019theoretically} which is a variant. This  differs from adversarial fine-tuning introduced in Section \ref{sec:methodology} since it trains the entire network instead of just the final layer. Since \alg\ uses larger models to train a smaller student network, we also compare it against the performance of distilling directly from an ERM trained teacher model. The teacher in all cases is ResNet-152.

\subsection{Results}
We compare the OOD accuracy of \alg against baselines  in Table~\ref{tab:results}.
We notice that our method provides significant improvements over standard ERM across all datasets and model sizes 
For example, on OfficeHome, we show gains of almost 5\% over ERM on all model sizes. We also note that smaller models trained with \alg outperform ERM trained larger models; ResNet-34 trained with \alg is close in performance to ResNet-152 on an average, while ResNet-50 can beat it. \\
We also notice that our method outperforms standard logit distillation on all benchmark datasets. This demonstrates that our method leverages the information provided by larger models in a more efficient manner. 
Furthermore, the KL-regularization of teachers in DAFT helps improve the transfer, as we demonstrate in the ablation experiments (sec~\ref{sec:ablations}). \\ 
We also observe that the method from \cite{zhang2021subnetwork} works only slightly better than ERM -- on the OfficeHome dataset, the accuracies are 58.5\%, 61.7\%, 67.3\%, 67.7\% and 69.5\% for ResNets 18-152 respectively. Their performance improvements are within the standard deviation of ERM trained models, and hence we do not compare against the method on the rest of DomainBed.
We also notice that \alg is able to outperform standard baselines by larger margins for datasets like TerraIncognita where the domains are significantly different from ImageNet. This means that features learnt from ImageNet pretraining would be less useful in this scenario. The better performance of DAFT on this dataset implies that it is able to transfer generalizable features better. Further, we perform experiments to verify the gains of \alg without using ImageNet pre-trained networks on datasets from the WILDS~\cite{koh2021wilds} benchmark in the appendix.

\vspace*{-10pt}

\subsection{Ablations}
\label{sec:ablations}

\begin{table}[!t]

\begin{center}
\begin{small}
\begin{sc}
\begin{tabular}{c|cccc|c}
\toprule
Model & Algorithm &  PACS  & VLCS & OfficeHome & Avg\\
\midrule
\multirow{5}{*}{ResNet-101} & ERM & 85.0\std{0.0} & 76.9\std{0.4} & 67.6\std{0.5} & 76.5\\
 & AT & 72.6\std{0.1} & 75.9\std{0.4} & 67.5\std{0.4} & 72.0\\
 & AF & \textbf{86.4\std{0.1}} & \textbf{77.9\std{0.1}} & \textbf{69.6\std{0.2}} & \textbf{77.9}\\
 & TRADES & 83.7\std{0.5} & 76.3\std{0.4} & 68.0\std{0.3} & 76.0\\
 & AF+$\cL_{smooth}$ & \textbf{86.5\std{0.1}} & \textbf{77.6\std{0.3}} & \textbf{69.5\std{0.3}} & \textbf{77.9}\\
\midrule
\multirow{5}{*}{ResNet-152} & ERM & 87.0\std{0.4} & 79.2\std{0.1} & 69.0\std{0.5} & 78.4\\
 & AT & 87.1\std{0.1} & 78.8\std{0.1} & 69.6\std{0.3} & 78.5\\
 & AF & \textbf{88.3\std{0.1}} & \textbf{80.4\std{0.0}} & \textbf{70.9\std{0.2}} & \textbf{79.9}\\
 & TRADES & 87.3\std{0.1} & 78.8\std{0.1} & 69.7\std{0.1} & 78.6\\
 & AF+$\cL_{smooth}$ & \textbf{88.4\std{0.4}} & \textbf{80.4\std{0.0}} & \textbf{70.9\std{0.3}} & \textbf{79.9}\\

\bottomrule
\end{tabular}
\end{sc}
\end{small}\vspace*{2pt}
\caption{\textbf{Effect of adversarial finetuning (AF)}: Accuracy achieved by ERM, adversarial training (AT) and adversarial finetuning (AF) for different architectures. Note that AF performs better than ERM, while AT is often worse than or similar to ERM due to poor in-domain accuracy of AT.
\label{tab:af_teacher}}

\end{center}
\vskip -0.1in\vspace*{-5pt}
\end{table}

\vspace*{-5pt}
\paragraph{Does adversarial finetuning work?}
To study the effect of our teacher training paradigms, we compare performance of various teacher models on the PACS, VLCS and OfficeHome dataset in Table~\ref{tab:af_teacher}. 
We show that it is much better to pre-train a model and finetune the final layer adversarially (AF), rather than training the full model adversarially (AT). Note that AT is competitive to ERM only on the OfficeHome dataset, since there is a high inter-domain similarity in three of the four domains of this dataset, and the images are also similar to ImageNet, on which the models were originally pretrained. This is consistent with the findings of \cite{yi2021improved}.

\begin{figure}[!t]
\includegraphics[width=\linewidth]{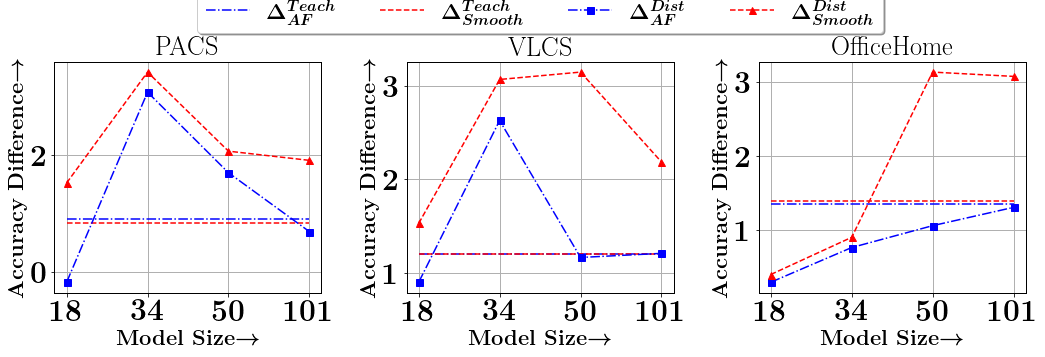}\vspace*{-3pt}
\caption{\textbf{Importance of smoothness term $\cL_{\textrm
{smooth}}$ in teacher training for student performance}: $(\Delta^\textrm{Dist}_{\textrm{Smooth}})$ and $(\Delta^\textrm{Dist}_{\textrm{AF}})$ denote accuracy increment for student models  compared to std distillation, when distilled from a teacher with and without the smoothness term, respectively. Accuracy improvements for teacher over ERM are $\Delta^\textrm{Teach}_{\textrm{AF}}$ and $\Delta^\textrm{Teach}_{\textrm{Smooth}}$ when trained with and without smoothness term,  respectively. Clearly, training the teacher with the smoothness term $\cL_\textrm{Smooth}$ leads to accuracy improvements for students over and above  improvements to the teacher accuracy.
\label{fig:smoothness}
\vspace*{2pt}}

\vskip -0.1in
\end{figure}

\noindent\textbf{Effect of $\cL_{smooth}$ on distillation}:
To verify the effect of using teachers trained with $\cL_{\textrm{smooth}}$, we present results on three datasets in Fig~\ref{fig:smoothness}. For each split, we compute the average gain in OOD accuracy for the teacher (which is a ResNet-152) when trained with adversarial finetuning with ($\Delta_{\cL_{\textrm{smooth}}}^{\textrm{Teach}}$) and without ($\Delta_{\textrm{AF}}^{\textrm{Teach}}$) $\cL_{\textrm{smooth}}$. We then compare the gains over standard distillation observed in students distilled from these teachers ($\Delta_{\cL_{\textrm{smooth}}}^{\textrm{Dist}}$ and $\Delta_{\textrm{AF}}^{\textrm{Dist}}$ respectively). Note that gain here refers to the difference in the OOD accuracy of the modified teacher (resp. distilled student of the modified teacher) model over a standard ERM (resp. distilled student of a standard teacher) model.
We observe that students distilled from a teacher trained with $\cL_{\textrm{smooth}}$ obtain similar or even better accuracy gains compared to those achieved by the teacher. In contrast, students of teachers trained without this term do not even consistently achieve similar accuracy gains as their teachers.

\begin{table}[!ht]

\begin{center}
\begin{small}
\begin{sc}
\begin{tabular}{c|cccc|c}
\toprule
Model & Algorithm                                                         & PACS                                  & VLCS 
                       & OfficeHome & Avg\\
\midrule
\multirow{2}{*}{ResNet-18} & DAFTSingle & 82.39\std{0.3} & 77.72\std{0.3} & 62.73\std{0.7} & 74.44\\
 & DAFT & 84.65\std{1.1} & 78.16\std{0.1} & 63.20\std{0.2}  & 75.18\\
\midrule
\multirow{2}{*}{ResNet-34} & DAFTSingle & 86.90\std{0.4} & 78.86\std{0.5} & 67.09\std{0.4} & 77.62\\
 & DAFT & 87.43\std{0.3} & 79.08\std{0.9} & 67.16\std{0.5} & 77.89\\
\midrule
\multirow{2}{*}{ResNet-50} & DAFTSingle & 87.66\std{0.2} & 78.73\std{0.5} & 69.34\std{0.4} & 78.58\\
 & DAFT & 87.86\std{0.2} & 79.64\std{0.4} & 70.85\std{0.3} & 79.45\\
\midrule
\multirow{2}{*}{ResNet-101} & DAFTSingle & 88.24\std{1.0} & 78.67\std{0.6} & 70.66\std{0.3} & 79.19\\
 & DAFT & 88.78\std{0.5} & 79.13\std{0.5} & 72.16\std{0.8} & 80.03\\
\bottomrule
\end{tabular}
\vspace*{2pt}
\caption{\textbf{\alg vs \algsingle}: Accuracies of \alg and \algsingle show that using multiple teachers for distillation as in \alg leads to significant improvements over using a single 
adversarially finetuned \color{black} teacher as in \algsingle.
\label{tab:office_multi}
\vspace*{-10pt}}
\end{sc}
\end{small}
\end{center}
\end{table}

\noindent\textbf{Effect of multiple teachers}:
We compare the OOD performance of student networks when trained with single or multiple teachers, i.e., \algsingle and \alg respectively, in Table~\ref{tab:office_multi}. We note that using multiple teachers consistently produce better students on average.
\vspace*{-5pt}

\section{Conclusion}
\label{sec:conclusion}\vspace*{-6pt}
\textbf{Summary}: In this paper, we considered the problem of out of distribution (OOD) generalization, where we are given training examples from a source distribution and are required to output a model which will be evaluated on test examples sampled from a different target distribution. We first observed that the non-robustness of standard trained models on OOD data is primarily due to a non-robust combination of learned features in the final linear layer and that the features themselves are capable of obtaining high OOD accuracy. Inspired by this observation, we designed adversarial finetuning (AF) which first trains the model using standard training and then finetunes the final linear layer using adversarial training. 

Motivated by the in-domain accuracy improvements obtained by distillation in prior works, we attempted to train a student model by distilling a teacher model that is trained by AF. However, we observed that standard distillation does not yield large improvements for AF trained teachers. We identified the reason for this to be the instability of logit values around the input and to address this, we incorporated an additional loss term in AF to encourage the logit values of teacher to be smooth. Finally, to tackle the suboptimal in-domain accuracy of AF trained teacher, we distilled  both standard trained and AF trained teachers into the student giving our final algorithm \alg. 

On five benchmark datasets, with diverse kinds of distribution shifts, we showed that \alg provides significantly higher OOD accuracy when compared to ERM as well as baselines like adversarial training. We also presented ablation studies showing the importance of various components of \alg. 

\textbf{Limitations \& Future work}: While our experiments cover a diverse array of distribution shifts, and show that \alg performs well, there may be other forms of distribution shift requiring further algorithmic ideas. Another avenue for future work is devising the optimal way of performing AF -- in this work, we only consider finetuning the final linear layer but have not explored if this can be further improved by finetuning last few layers or  other subsets of parameters. Finally, models that are fragile to OOD shifts and depend on spurious correlation  can significantly amplify biases in data. So, further investigation of \alg for mitigating biases in data is highly interesting. 
 \clearpage

\bibliographystyle{unsrt}
\bibliography{references}

\clearpage
\newpage
\appendix
\section{Experimental Setup}
\label{app:Problem Definition}
{We use the same problem setting as Domainbed benchmark. We consider samples $(x_1,y_1),...,(x_n,y_n)$ drawn from a training distribution $\mathcal{D}_s$. During test time, samples are drawn from a test distribution $\mathcal{T}$. Following Domainbed benchmark, we assume access to a pool of validation samples that are aggregated from multiple target domains. We cannot use these samples to train the model.\\
To further illustrate,  let’s consider the  example of the OfficeHome dataset. There are four sub-domains of the dataset - Art, Clipart, Photo and Product. For each of these sub-domains, the data is divided into a 80-20 split. Since there are four different sub-domains of this dataset, we essentially consider 4 different instances of the OOD generalization problem, wherein each problem considers one sub-domain as the test distribution, and the union of the remaining three as the train distribution. Both training and testing are always done on the 80\% splits of the respective sub-domains. We then report the average test accuracy on all four instances of the problem for each method. We run this experiment 5 times across random restarts with the same hyper-parameters, and report the mean and standard deviations of the average test accuracy for each algorithm. \\
For hyperparameter selection, we use the leave one out setting mentioned in the Domainbed work \cite{gulrajani2020search} That is, given the four sub-domains of OfficeHome, we first sample $k$ hyper-parameter settings for an algorithm. For each of the $k$ hyper-parameter settings, we then train four models using the algorithm, leaving out one of the domains each time, and noting the accuracy of the model on the 20\% split of the held out domain. The average of the four held-out accuracies is the validation accuracy of the hyper-parameter choice. We then choose the hyper-parameter setting which maximizes this validation accuracy.}
\color{black}
\label{app:params}
\subsection{Hyperparameters}
We use the Adam optimizer for all our experiments, with a batch size of 64. We run ERM, adversarial training, and distillation for 10000 steps each, while fine-tuning is run for 5000 steps. We tune the following hyper-parameters for our methods and baselines - 
\begin{itemize}
\item Learning Rate - Selected from the range $[10^{-6}, 10^{-3}]$
\item Norm of adversarial perturbation $\epsilon$ - Selected from the range $[0.05, 0.5]$
\item Number of adversarial perturbation steps $k$ - An integer selected from the range $[3,7]$ 
\item LR for PGD - Selected from the range $[10^{-3}, 10^{-1}]$
\item Distillation temperature $\tau$ - An integer selected from the range $[2,8]$
\item Weight $\alpha$  for $\cL_{smooth}$ - Selected from the range $[10^{-6}, 10^{-2}]$
\end{itemize}

The hyperparameters were tuned using random search over the intervals, with $32$ configurations being considered for each algorithm.

\subsection{Datasets}
The data can be downloaded using the DomainBed \href{https://github.com/facebookresearch/DomainBed/tree/main/domainbed}{repo}
\subsection{Hardware Setup}
We conducted all experiments on a single A100 GPU. The experimental code was using the DomainBed framework, in PyTorch.
\section{Additional results}
\subsection{Statistical analysis of main results}
We conduct paired t-tests of the results reported in table~\ref{tab:results}. In particular, we report the p-values of the the paired-t test between the results obtained by DAFT and Distillation in table~\ref{tab:ttest}. We boldface a value in table~\ref{tab:results} if the corresponding p-value is less than 0.05.
\begin{table}[!ht]

\begin{center}
\begin{small}
\begin{sc}
\begin{tabular}{c|ccccc}
\toprule
Model & PACS & VLCS & OfficeHome & DomainNet & TerraIncognita \\
\midrule
ResNet-18 & 0.132 & 0.011 & 0.020 & 0.003 & 0.016 \\
ResNet-34 & 0.023 & 0.021 & 0.031 & 0.008 & 0.015 \\
ResNet-50 & 0.011 & 0.018 & 0.010 & 0.015 & 0.001 \\
ResNet-101 & 0.034 & 0.024 & 0.025 & 0.443 & 0.002 \\
ResNet-152 & 0.117 & 0.449 & 0.212 & 0.077 & 0.315 \\
\bottomrule
\end{tabular}
\vspace*{2pt}
\caption{p-values of the paired t-test between the performance of DAFT and Distillation .
\label{tab:ttest}
\vspace*{-10pt}}
\end{sc}
\end{small}
\end{center}
\end{table}
\color{black}
\label{app:additional}
\subsection{Results with the MobileNet architecture}
We try out our method on mobilenet class of models (cite) and demonstrate that the observations for our method generalize form resnet models to mobilenet clas sof models as well. We perform experiments with the MobileNet \cite{howard2019searching} architecture where the teacher is a ResNet-152. The results are listed in table~\ref{tab:mobnet}.
\begin{table}[!ht]

\begin{center}
\begin{small}
\begin{sc}
\begin{tabular}{c|ccccc}
\toprule
Model & Algorithm                                                         & PACS                                  & VLCS 
                       & OfficeHome & TerraIncognita\\
\midrule
\multirow{3}{*}{MobileNetv3-Small} & ERM & 80.2 & 76.4 & 53.6 & 37.2\\
 & Distillation & 81.3 & 77.2 & 57.6 & 42.5 \\
 & DAFT & 81.6 & 77.6 & 58.5 & 44.4 \\
\midrule
\multirow{3}{*}{MobileNetv3-Large} & ERM & 85.9 & 79.8 & 63.1 & 47.8 \\
 & Distillation & 86.9 & 81.1 & 66.5 & 51.1 \\
 & DAFT & 87.5 & 82.1 & 67.2 & 53.8 \\
\bottomrule
\end{tabular}
\vspace*{2pt}
\caption{Comparison on the MobileNet family of architectures.
\label{tab:mobnet}
\vspace*{-10pt}}
\end{sc}
\end{small}
\end{center}
\end{table}
\color{black}
\subsection{Results on WILDS benchmark}
We perform experiments on the WILDS benchmark~\cite{koh2021wilds} with non-ImageNet pre-trained models to verify the efficacy of our approach in settings where pre-training on large datasets is not possible. We present results on iWildCams in Table~\ref{tab:iwildcam}. We notice that DAFT consistently gives gains over all the baselines.

\begin{table}[!ht]

\begin{center}
\begin{small}
\begin{sc}
\begin{tabular}{cccccc}
\toprule
Model      & ERM & Dist & Adv & Trades & \alg \\
\midrule
RN-200 & 56.1 \std{0.4}                    & 56.7 \std{0.4}           &          51.2 \std{1.2}                 &   53.0 \std{1.2}     & \textbf{58.7 \std{0.6} }                    \\
RN-101 & 50.9 \std{0.3}                    & 55.8 \std{1.5}                               & 47.6 \std{1.0}                         & 50.2 \std{0.8}                       & \textbf{58.1 \std{0.4}}                     \\
RN-50  & 49.1 \std{0.5}                    & 55.4 \std {1.1}                             & 45.4 \std{1.1}                           & 47.9 \std{1.7}                       & \textbf{57.2 \std{0.5}}                    \\
RN-34  & 47.6 \std{0.9}                    & 52.5 \std{0.9}                             & 42.3 \std{1.7}                         & 45.5 \std{1.6}                       & \textbf{55.2 \std{1.5}}                     \\
RN-18  & 44.9 \std{0.6}                    & 50.7 \std{1.3}                             & 41.2 \std{1.8}                           & 42.7 \std{1.1}                       & \textbf{54.3 \std{1.3}}    \\                

\bottomrule
\end{tabular}
\caption{iWildCam dataset: OOD accuracy for different student architectures. RN refers to the ResNet architecture family. For all the considered RN models, \alg is significantly more accurate than ERM and DIST, while vanilla ADV/TRADES training leads to worse OOD accuracy than ERM. \vspace*{-10pt}
\label{tab:iwildcam}}
\end{sc}
\end{small}
\end{center}
\end{table}

\subsection{Detailed results on DomainBed}
We present the domain-wise accuracies of DAFT for each of the datasets in DomainBed in tables \ref{tab:pacs}-\ref{tab:terra_incognita}.

\begin{table}[]
\begin{center}
\begin{small}
\begin{sc}
\begin{tabular}{c|cccc|c}
\toprule
Model Size & A & C & P & S & Avg \\
\midrule
ResNet-18 & 82.5\std{2.0} & 82.7\std{1.8} & 93.1\std{2.1} & 80.8\std{0.1} & 84.8\\
ResNet-34 & 87.4\std{0.9} & 83.3\std{0.5} & 94.4\std{0.6} & 84.4\std{0.4} & 87.4\\
ResNet-50 & 88.2\std{1.4} & 84.3\std{1.1} & 94.6\std{0.8} & 84.9\std{0.5} & 88.0\\
ResNet-101 & 90.5\std{0.0} & 84.1\std{2.0} & 96.4\std{0.0} & 84.2\std{0.0} & 88.8\\
ResNet-152 & 88.1\std{1.0} & 84.3\std{1.3} & 95.8\std{0.9} & 86.3\std{1.1} & 88.6\\
\bottomrule
\end{tabular}\vspace*{2pt}
\caption{OOD accuracy of \alg on various domains of PACS different ResNet (RN) architectures.
\label{tab:pacs}
}
\end{sc}
\end{small}
\end{center}

\end{table}

\begin{table}[]
\begin{center}
\begin{small}
\begin{sc}
\begin{tabular}{c|cccc|c}
\toprule
Model Size                           & C & L & S & V & Avg \\
\midrule
ResNet-18 & 98.4\std{0.2} & 67.4\std{0.5} & 71.5\std{0.7} & 75.3\std{0.4} & 78.2\\
ResNet-34 & 98.7\std{0.4} & 66.5\std{1.7} & 73.3\std{1.4} & 77.9\std{0.6} & 79.1\\
ResNet-50 & 98.7\std{0.4} & 68.9\std{0.8} & 73.7\std{0.6} & 78.5\std{0.4} & 80.0\\
ResNet-101 & 98.6\std{0.0} & 69.6\std{1.8} & 73.6\std{0.1} & 74.6\std{0.1} & 79.1\\
ResNet-152 & 98.4\std{0.5} & 69.9\std{1.3} & 73.6\std{0.2} & 80.7\std{2.2} & 80.7\\

\bottomrule
\end{tabular}\vspace*{2pt}
\caption{OOD accuracy of \alg on various domains of VLCS different ResNet (RN) architectures.
\label{tab:vlcs}
}
\end{sc}
\end{small}
\end{center}

\end{table}

\begin{table}[]
\begin{center}
\begin{small}
\begin{sc}
\begin{tabular}{c|cccc|c}
\toprule
Model Size                  & Art & Clipart & Product & Real  & Avg \\
\midrule
ResNet-18 & 56.7\std{1.3} & 51.7\std{0.2} & 69.9\std{0.4} & 74.5\std{0.3} & 63.2\\
ResNet-34 & 61.1\std{1.3} & 56.0\std{1.1} & 74.6\std{0.3} & 77.3\std{0.3} & 67.2\\
ResNet-50 & 67.6\std{0.6} & 57.4\std{0.5} & 77.8\std{0.4} & 81.0\std{0.2} & 71.0\\
ResNet-101 & 68.9\std{0.0} & 59.5\std{0.1} & 78.4\std{0.3} & 81.9\std{0.1} & 72.2\\
ResNet-152 & 71.2\std{0.3} & 59.6\std{0.3} & 79.1\std{0.5} & 82.3\std{0.2} & 73.1\\
\bottomrule
\end{tabular}\vspace*{2pt}
\caption{OOD accuracy of \alg on various domains of OfficeHome different ResNet (RN) architectures.
\label{tab:office}
}
\end{sc}
\end{small}
\end{center}

\end{table}

\begin{table}[]
\begin{center}
\begin{small}
\begin{sc}
\begin{tabular}{c|cccc|c}
\toprule
Model Size                  & L100 & L38 & L48 & L46 & Avg \\
\midrule

ResNet-18 & 59.7\std{2.0} & 46.7\std{1.5} & 55.5\std{1.7} & 39.1\std{0.8} & 50.3\\
ResNet-34 & 61.1\std{2.3} & 48.0\std{1.2} & 58.0\std{0.6} & 38.7\std{1.2} & 51.5\\
ResNet-50 & 61.1\std{1.1} & 49.4\std{1.2} & 59.9\std{0.8} & 40.7\std{0.3} & 52.8\\
ResNet-101 & 62.2\std{1.4} & 50.1\std{1.7} & 60.7\std{0.7} & 43.2\std{0.8} & 54.0\\
ResNet-152 & 65.8\std{0.8} & 53.2\std{0.9} & 61.7\std{1.1} & 42.9\std{1.2} & 55.9\\
\bottomrule
\end{tabular}\vspace*{2pt}
\caption{OOD accuracy of \alg on various domains of TerraIncognita different ResNet (RN) architectures.
\label{tab:terra_incognita}
}
\end{sc}
\end{small}
\end{center}

\end{table}

\begin{table}[]
\begin{center}
\begin{small}
\begin{sc}
\begin{tabular}{c|cccccc|c}
\toprule
Model Size        & R & C & I & Q &P & S & Avg \\
\midrule
ResNet-18 & 46.9\std{0.1} & 54.1\std{0.3} & 17.4\std{0.5} & 11.2\std{0.5} & 44.3\std{0.3} & 44.7\std{0.1} & 36.4\\
ResNet-34 & 50.8\std{1.8} & 55.8\std{0.4} & 19.4\std{0.4} & 12.8\std{0.2} & 46.5\std{0.2} & 45.8\std{1.2} & 38.5\\
ResNet-50 & 61.0\std{0.0} & 57.6\std{0.0} & 22.6\std{0.6} & 15.7\std{0.2} & 48.9\std{1.6} & 49.2\std{0.6} & 42.5\\
ResNet-101 & 63.8\std{0.3} & 59.7\std{0.2} & 23.5\std{0.3} & 16.4\std{0.3} & 48.5\std{0.2} & 49.8\std{0.1} & 43.6\\

\bottomrule
\end{tabular}\vspace*{2pt}
\caption{OOD accuracy of \alg on various domains of DomainNet different ResNet (RN) architectures.
\label{tab:domainnet}
}
\end{sc}
\end{small}
\end{center}

\end{table}

\subsection{Results using Oracle strategy for hyper-parameter selection}
Gulrajani et al. \cite{gulrajani2020search} note that using a hold-out validation set from the target domain can be a strategy for hyperparameter tuning in domain generalization. In table~\ref{tab:oracle_results}, we present the results obtained by using this on 4 datasets of DomainBed. We note that DAFT still out-performs other baselines in this case, and is considerably better than the results reported in Table-1 of the main paper.

\begin{table}[]
\begin{center}
\begin{small}
\begin{sc}
\resizebox{\linewidth}{!}{
\begin{tabular}{c|cccccc|c}
\toprule
Model Size                  & Method               & PACS & VLCS & OfficeHome & TerraIncognita & DomainNet & Avg \\
\midrule
\multirow{5}{*}{ResNet-18} & ERM & 84.9\std{0.4} & 76.8\std{0.2} & 60.3\std{0.3} & 48.4\std{0.8} & 32.2\std{0.8} & 60.6\\
 & AT & 81.6\std{0.5} & 69.4\std{1.0} & 61.8\std{0.1} & 57.9\std{0.0} & 32.2\std{0.2} & 60.6\\
 & TRADES & 80.8\std{0.7} & 72.0\std{1.0} & 57.8\std{1.0} & 41.6\std{0.3} & 32.1\std{1.4} & 56.9\\
 & Distillation & 85.2\std{0.3} & 78.7\std{0.4} & 65.0\std{0.1} & 52.1\std{0.0} & 35.2\std{1.1} & 63.3\\
 & DAFT & 86.8\std{0.2} & 79.8\std{0.3} & 65.1\std{0.1} & 52.9\std{0.4} & 38.2\std{0.7} & 64.6\\
\midrule
\multirow{5}{*}{ResNet-34} & ERM & 88.2\std{0.4} & 79.6\std{0.4} & 64.7\std{0.1} & 50.9\std{1.0} & 33.6\std{0.5} & 63.4\\
 & AT & 84.2\std{0.2} & 75.3\std{0.2} & 65.2\std{0.3} & 43.1\std{0.8} & 32.5\std{1.2} & 60.1\\
 & TRADES & 84.0\std{1.3} & 73.5\std{0.1} & 62.9\std{0.4} & 43.7\std{0.8} & 33.3\std{1.1} & 59.5\\
 & Distillation & 88.7\std{0.2} & 79.9\std{0.1} & 68.8\std{0.2} & 54.2\std{0.4} & 38.9\std{0.1} & 66.1\\
 & DAFT & 89.3\std{0.1} & 81.3\std{0.3} & 68.7\std{0.1} & 54.8\std{0.3} & 39.7\std{1.1} & 66.8\\
\midrule
\multirow{5}{*}{ResNet-50} & ERM & 88.7\std{0.2} & 79.4\std{0.5} & 68.4\std{0.6} & 53.9\std{0.8} & 42.1\std{0.0} & 66.5\\
 & AT & 84.7\std{1.4} & 74.1\std{0.4} & 69.4\std{0.1} & 46.2\std{1.4} & 41.6\std{1.4} & 63.2\\
 & TRADES & 84.0\std{0.4} & 74.8\std{1.2} & 67.1\std{1.2} & 47.5\std{1.4} & 41.7\std{1.3} & 63.0\\
 & Distillation & 89.6\std{0.2} & 80.3\std{0.3} & 70.7\std{0.1} & 55.4\std{0.4} & 43.5\std{1.2} & 67.9\\
 & DAFT & 90.3\std{0.4} & 81.6\std{0.5} & 72.5\std{0.3} & 56.6\std{0.4} & 45.0\std{0.7} & 69.2\\
\midrule
\multirow{5}{*}{ResNet-101} & ERM & 88.1\std{0.0} & 79.1\std{0.0} & 68.7\std{0.0} & 52.1\std{0.0} & 44.2\std{1.0} & 66.4\\
 & AT & 73.6\std{0.7} & 76.4\std{1.4} & 71.8\std{0.1} & 49.3\std{0.8} & 44.4\std{0.9} & 63.1\\
 & TRADES & 84.9\std{1.3} & 78.8\std{0.3} & 69.4\std{0.9} & 51.4\std{0.6} & 44.2\std{1.2} & 65.7\\
 & Distillation & 90.8\std{0.2} & 80.5\std{0.1} & 71.8\std{0.1} & 55.6\std{0.4} & 45.1\std{1.0} & 68.8\\
 & DAFT & 91.5\std{0.0} & 82.2\std{0.1} & 73.0\std{0.2} & 56.5\std{0.2} & 45.5\std{0.7} & 69.8\\
\midrule
\multirow{5}{*}{ResNet-152} & ERM & 88.4\std{0.0} & 79.2\std{0.0} & 69.8\std{0.3} & 54.4\std{0.0} & 43.8\std{0.9} & 67.1\\
 & AT & 89.5\std{0.7} & 80.3\std{0.6} & 72.1\std{0.2} & 50.6\std{0.2} & 44.4\std{1.4} & 67.4\\
 & TRADES & 89.3\std{0.8} & 81.1\std{0.9} & 70.4\std{0.1} & 51.2\std{1.1} & 43.5\std{1.4} & 67.1\\
 & Distillation & 90.7\std{0.2} & 82.3\std{0.2} & 72.9\std{0.3} & 56.5\std{0.2} & 45.2\std{1.0} & 69.5\\
 & DAFT & 91.3\std{0.3} & 82.4\std{0.2} & 73.2\std{0.1} & 57.8\std{0.4} & 45.0\std{0.9} & 69.9\\

\bottomrule
\end{tabular}}\vspace*{2pt}
\caption{OOD accuracy on DomainBed using the Oracle selection strategy for hyperparameter tuning.
\label{tab:oracle_results}
}
\end{sc}
\end{small}
\end{center}

\end{table}

\section{Verifying our Design Choices}

\subsection{Perturbing features instead of the input}
We also experiment with a variant of adversarial fine-tuning where we perform perturbations in the feature space of the model rather than the input space. The comparison with adversarial finetuning on four datasets is reported in table~\ref{tab:af_last}. We notice that the difference in the performance obtained is not consistent across datasets or sizes. While the performance of finetuning in feature space is slightly more for ImageNet pre-trained models, we notice that the range over which $\epsilon$ needs to be fine-tuned is larger for this variant, and the obtained best $\epsilon$ differs quite a bit between different models. On the contrary, for input space perturbations, using the same $\epsilon$ across models does not degrade performance noticeably. 

Input perturbations can potentially have an edge over feature space perturbations when the training data diversity is limited, leading to feature replication. In such a case, perturbing the relevant input pixels can perturb all the replicated features, while feature space perturbations need to individually perturb every replicated feature within the given perturbation budget. This is indeed the case when we do not use ImageNet pretraining, as seen in the last column of table~\ref{tab:af_last}.
\begin{table}[!th]
\begin{center}
\begin{small}
\begin{sc}
\begin{tabular}{c|ccccc}
\toprule
Model Size                  & Method               & PACS & VLCS & OfficeHome & FMoW \\
\midrule
\multirow{2}{*}{ResNet-101} & AF & 86.4 & 77.9 & 69.6 & 51.8\\
& AFLast & 87.0 & 77.1 & 70.3 &  49.6\\
\midrule
\multirow{2}{*}{ResNet-152} & AF & 88.3 & 80.4 & 70.9 & 55.0 \\
& AFLast & 88.9 & 80.1 & 71.1 & 54.7 \\

\bottomrule
\end{tabular}\vspace*{2pt}
\caption{Comparison between input perturbations and perturbations in the feature space.
\label{tab:af_last}
}
\end{sc}
\end{small}
\end{center}

\end{table}

\subsection{Fine-tuning multiple layers}
In table~\ref{tab:af_multi}, we fine-tune the last three layers instead of just the final layer (AFMulti). We find that this leads to slightly improved performance on two datasets, and slightly degraded performance on one. 
\begin{table}[!th]
\begin{center}
\begin{small}
\begin{sc}
\begin{tabular}{c|cccc}
\toprule
Model Size                  & Method               & PACS & VLCS & OfficeHome  \\
\midrule
\multirow{2}{*}{ResNet-101} & AF & 86.4 & 77.9 & 69.6 \\
& AFMulti & 87.1 & 77.7 & 70.5 \\

\bottomrule
\end{tabular}\vspace*{2pt}
\caption{
Effect of fine-tuning multiple layers.
\label{tab:af_multi}
}
\end{sc}
\end{small}
\end{center}

\end{table}

\subsection{Do adversarial perturbations lead to unstable logits?}
In order to verify the effect of adversarial finetuning with and without the $\cL_{smooth}$ loss, we compute the logits of an ERM trained model, an adversarially finetuned model and a KL-regularized finetuned model on the ``Clipart" split of the OfficeHome dataset. The training and finetuning data for the adversarially fine-tuned models were the ``Real", ``Product" and ``Painting" splits of the dataset, while the ERM trained model was trained on all datasets, i.e. it was also fine-tuned on the ``Clipart" split. We find that the mean Spearman rank correlation of the logits from the KL-regularized model with the ERM trained model is higher (0.528) when compared to that of the adversarially finetuned model (0.505). 
In table~\ref{tab:af_order}, we list the average precision@k (k between 1-5) of the logits from adversarially finetuned models with respect to those of the ERM model (trained on source domain and source+target domains resp.). Here prec@k is defined as $|topk\_predictions(model) \cap topk\_predictions(ERM Model)|/k$. As we can see, $\cL_{smooth}$ encourages the order of logits to be maintained (w.r.t. ERM model trained on source domain), while also being more similar to that of the ERM model trained on all domains, although its target accuracy is similar to the AF model.

\begin{table}[!th]
\begin{center}
\begin{small}
\begin{sc}

\begin{tabular}{c|c|ccccc|c}
\toprule
Method & ERM Model                  & prec@1               & prec@2 & prec@3 & prec@4 & prec@5  & MAP\\
\midrule
AF with $\cL_{smooth}$ & Src+Tgt & 62.5\% & 51.5\% & 48.9\% & 47.9\% & 47.5\% & 69.9\%                          \\
AF & Src+Tgt & 62.1\% & 49.4\% & 47.2\% & 46.9\% & 46.5\% & 67.4\%\\
\midrule
AF with $\cL_{smooth}$ & Src & 94.1\% & 63.1\% & 53.2\% & 49.7\% & 48.8\% & 72.6\%                          \\

AF & Src & 94.1\% & 59.3\% & 50.1\% & 48.1\% & 48.0\% & 69.4\%\\
\bottomrule
\end{tabular}\vspace*{2pt}
\caption{Comparison between teachers trained with and without $\cL_{smooth}$. We show the overlap in the order of the predictions here, and note that the OOD predictions of smooth teacher are better aligned with the ERM teacher.
\label{tab:af_order}
}
\end{sc}
\end{small}
\end{center}

\end{table}

\subsection{Additional results on Colored-FashionMNIST}
\begin{figure}[!ht]
\centering
\includegraphics[width=0.9\linewidth]{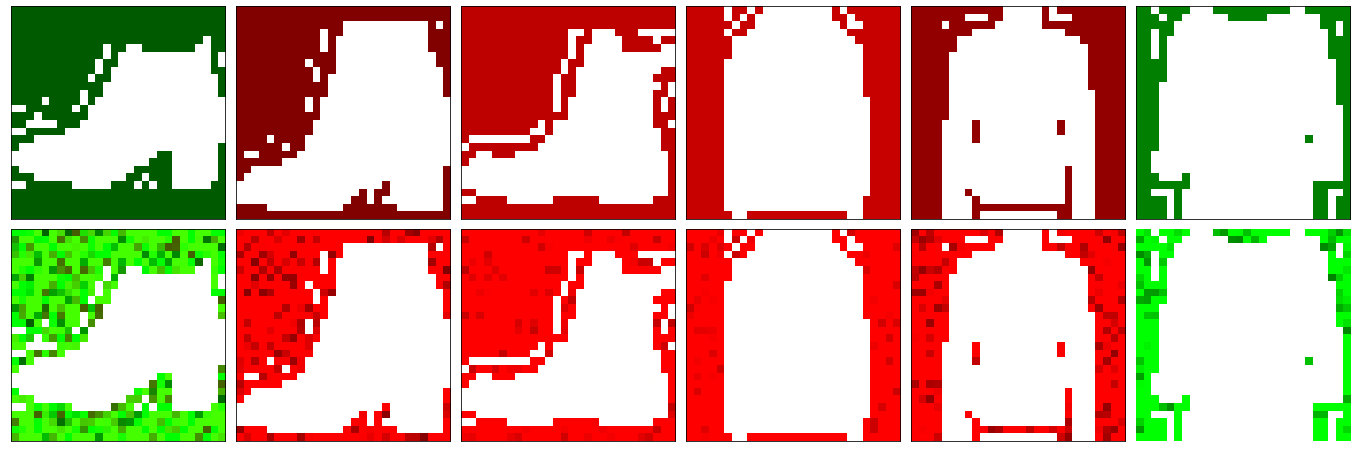}\vspace*{-5pt}
\caption{Sample images and their adversarially perturbed versions from Colored-FashionMNIST dataset. We notice that the color of the image is perturbed, while the shape remains constant.
\label{fig:fashion_mnist}} 
\end{figure}

In fig~\ref{fig:fashion_mnist}, we show examples of images from the FashionMNIST dataset, as well as the $l_2$-norm constrained adversarial perturbations. We choose the first three images of each label from the test set to perturb. We find that the perturbations mainly change the colour of the images. 
For each feature, we compute the relative average perturbation (RAP, i.e. $\mathbb{E}[ \frac{|f_i(x+\delta) - f_i(x)|}{|f_i(x)|}]$, where $\delta$ is the adversarial perturbation, and $f_i$ denotes the $i^{th}$ feature) when the input is perturbed adversarially. We call this RAP-input. We also compute the relative average perturbation when the adversarial perturbations are in the \textit{feature} space, denoted as RAP-feature. We notice that the maximum RAP-input for colour features is much more than that of shape features (11 v/s 0.4). This is expected since the adversarial perturbations only change the colour of the image. 
Hence, the model learns to weigh the colour features lesser while making predictions, since they change more while the label remains constant during adversarial fine-tuning\color{black}. Further, we also notice that there is a high correlation (0.842) between RAP-input and RAP-feature. In fact, RAP-feature also follows a similar trend, with the maximum RAP-feature for shape features being 0.3, while the maximum RAP-feature for colour features is 13. This means that fine-tuning the last layer with either feature perturbations or input perturbations would lead the model to similar classification weights. 

\end{document}